\documentclass{article} % For LaTeX2e
\usepackage{iclr2017_conference,times}
\usepackage{hyperref}
\usepackage{url}

\usepackage{natbib}
\usepackage{graphicx}
\usepackage{caption}
\usepackage{subcaption}
\usepackage{epstopdf}
\title{Binarized Neural Networks on the ImageNet Classification Task}

\author{Xundong Wu\\University of California, Los Angeles\\Los Angeles, CA, USA\\
\texttt{wuxundong@gmail.com}\\
\And
Yong Wu\\University of California, Los Angeles\\Los Angeles, CA, USA \\
\texttt{wuyong@ucla.edu}\\
\And
Yong Zhao\\Peking University, Shenzhen\\Shenzhen, China \\
\texttt{zhaoyong@pkusz.edu.cn}\\
}

\graphicspath{{./figures/}}
\begin{document}

\maketitle

\begin{abstract}
We trained Binarized Neural Networks (BNNs) on the high resolution ImageNet ILSVRC-2102 dataset classification task and achieved a good performance. With a moderate size network of 13 layers, we obtained top-5 classification accuracy rate of 84.1 percent on validation set through network distillation, much better than previous published results of 73.2\% on XNOR network and 69.1\% on binarized GoogleNET. We expect networks of better performance can be obtained by following our current strategies. We provide a detailed discussion and preliminary analysis on strategies used in the network training.
\end{abstract}

\section{Introduction}
\label{introduction}

Recent wave of deep learning research has brought the deep neural network to the frontier of AI applications. Current implementations of deep neural networks in wide variety of fields heavily rely on high computing power and energy hungry hardware such as GPUs, which limits implementations of those neural networks in embedded environment such as mobile phones or wearable hardware. Much effort has been put to reduce computing cost of deep neural networks, for example, \citep{han2015learning,chen2015compressing, jaderberg2014speeding}. 

Recently, \citep{rastegari2016xnor,courbariaux2016binarynet,courbariaux2015binaryconnect} have been trying to reduce synapse weights and/or intermediate signal representation to binary form. They have shown binarification of neural networks can dramatically reduce the computing cost \citep{rastegari2016xnor,courbariaux2016binarynet}, as well as the amount of memory needed for weight and intermediate results storage. Those approaches are especially useful for memory constrained environment such as FPGA implementations. 

In this work, we designed a network architecture based on Binarized Neural Networks (BNN), the neural network structure originally proposed by \citep{courbariaux2016binarynet}, and trained networks on the ILSVRC2012 image classification task \citep{krizhevsky2012imagenet}. 

We analyzed the behavior of weights adaptation during SGD training and use the observation to guide the setting of proper learning rates used in the network training. This leads to significant gain in the rate of convergence. 

In order to increase the recognition accuracy while maintains the very high efficiency of a BNN, a special new neural network architecture is proposed here. This network has more than usual channels than regular networks at early layers and normal layers are used through the rest layers. We obtained 80 percent top-5 accuracy on this network.

A further improvement of network performance is introduced through network distillation \citep{hinton2015distilling}. With a moderate size network an 84.1 percent top-5 accuracy on single crop validation set of ILSVRC2012 classification task is achieved, which is much better than previous published results on XNOR network \citep{rastegari2016xnor} and binarized version of GoogleNET \citep{hubara2016quantized}. We expect networks of better performance can be trained by following our current strategies.

\section{Results}
\label{results}

\subsection{First layer architecture}
In the original BNN layer designed for CIFAR-10 dataset as in \citep{courbariaux2016binarynet}, all layers of network are binary. For natural images, usually the first layer input only have 3 input channels, thus similar to \citep{rastegari2016xnor,hubara2016quantized}, we also use regular weight convolutional layer for the first layer of the network.

\begin{figure}
\begin{center}
	\begin{subfigure}{0.32\linewidth}
		{\includegraphics[width=\linewidth]{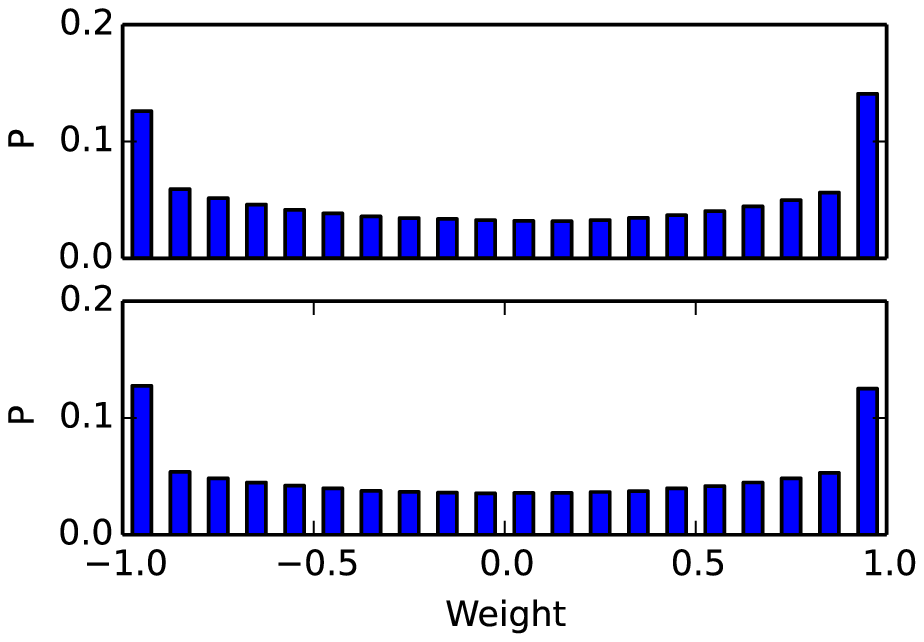}}
		\caption{Weight distribution, LR 0.001}
		\label{fig:f2a}
	\end{subfigure}
	\begin{subfigure}{0.32\linewidth}
		{\includegraphics[width=\linewidth]{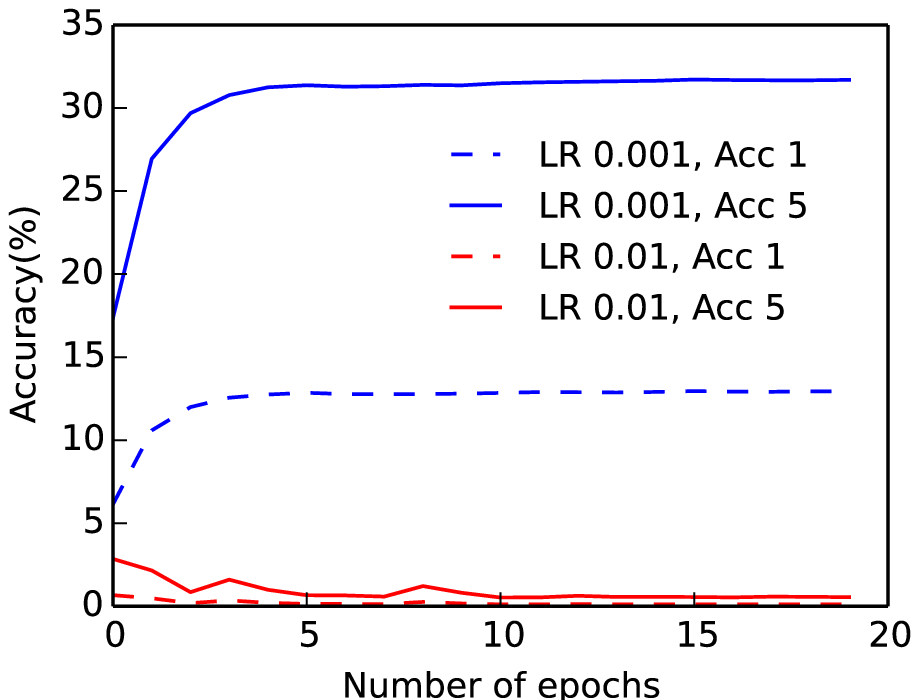}}
		\caption{Training curves}
		\label{fig:f2b}		
	\end{subfigure}
	\begin{subfigure}{0.32\linewidth}
		{\includegraphics[width=\linewidth]{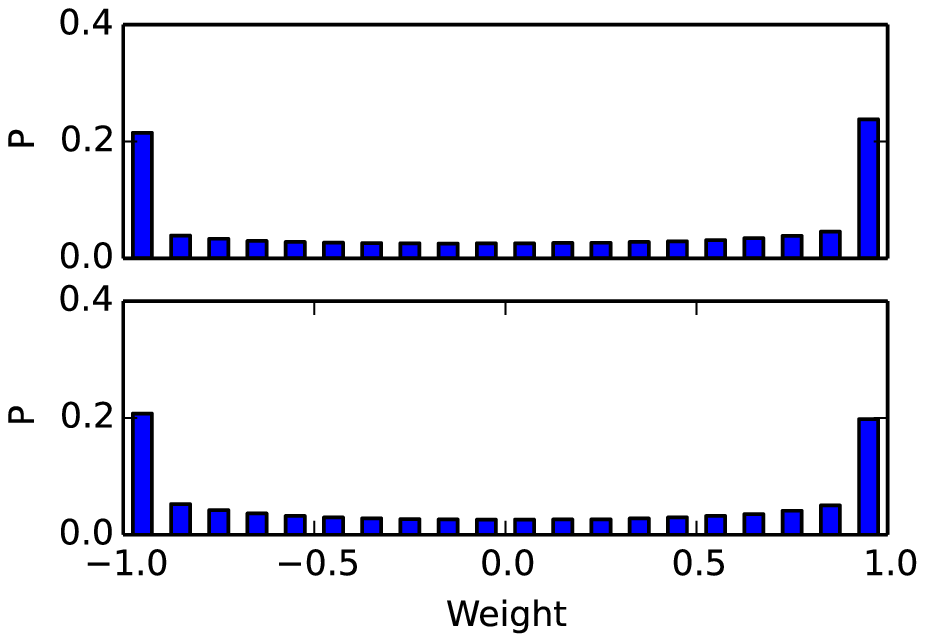}}
		\caption{Weight distribution, LR 0.01}
		\label{fig:f2c}		
	\end{subfigure}
\end{center}
\caption{Comparison of network convergence speed under two different learning rates.\\
A. and C. The distribution of synapse states after one epoch of training. Top: synapses from layer 2, Bottom: synapses from layer 6.\\B. Lower learing rate significantly accerelated convergence speed.\\Networks silmar to Alexnet are used here with batch size of 256 }\label{fig:f2}
\end{figure}

\subsection{Weight adaptation behavior analysis and learning rate selection}
In a typical BNN layer, at the training stage real-valued synaptic weights are used during SGD training. This extra states are removed at the stage of computing layer output. It is believed that real-valued weights are needed for smooth out the noise in SGD learning \citep{courbariaux2016binarynet}. In the field of computational neuroscience this is called binary synapses with internal/hidden states\citep{baldassi2007efficient,wu2009capacity}. Here we show indeed this is the case. When a high learning rate of 0.01 is used for training, after one epoch of training (we rescale learning rate of every binary layer as in \citep{courbariaux2016binarynet}), we observed most weight values stay around -1 and +1 (Figure \ref{fig:f2c}) because of weight clipping. Under such condition weights quickly jumped between edge values of -1 and +1 with little accumulation process, equivalently there are less hidden synapse states available to the training process. Correspondingly, a rather slow accuracy rate climbing process is observed (Figure \ref{fig:f2a}). When we use a lower learning rate of 0.001, we observed a weight distribution more uniform or concentrated around 0, indicating the weights are taking more steps to travel between -1 and +1 (Figure \ref{fig:f2b}) resemble the behavior of synapse age tags as in\citep{wu2009capacity} or hidden states in \citep{baldassi2007efficient}. Correspondingly, a much faster accuracy climbing was observed during the training process (Figure \ref{fig:f2a}).

\begin{figure}
\begin{center}
	\begin{subfigure}{0.38\linewidth}
		{\includegraphics[width=\linewidth]{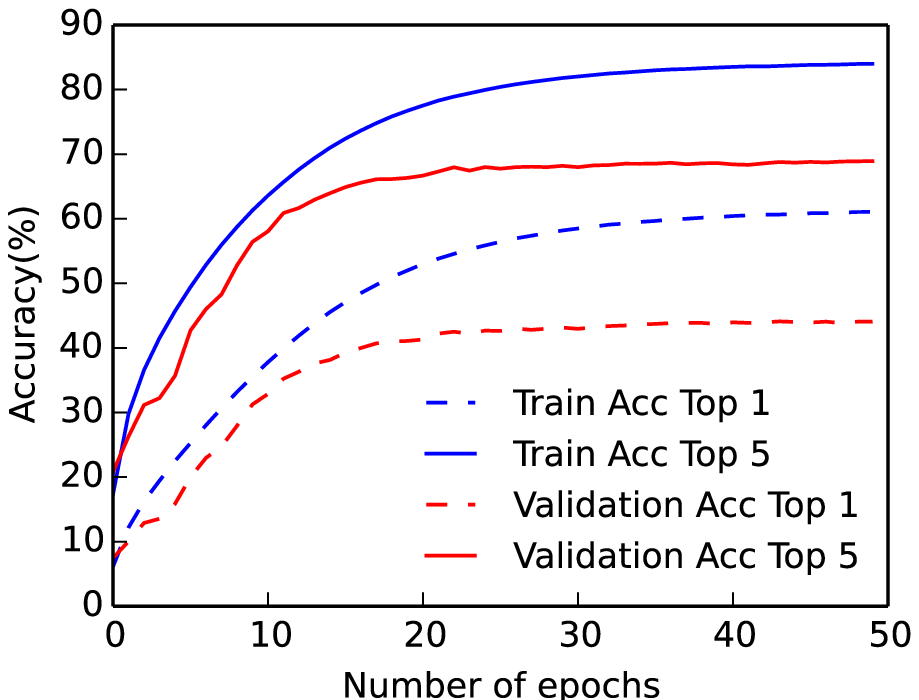}}
		\caption{Alexnet network}
		\label{fig:f3a}
	\end{subfigure}
	\begin{subfigure}{0.38\linewidth}
		{\includegraphics[width=\linewidth]{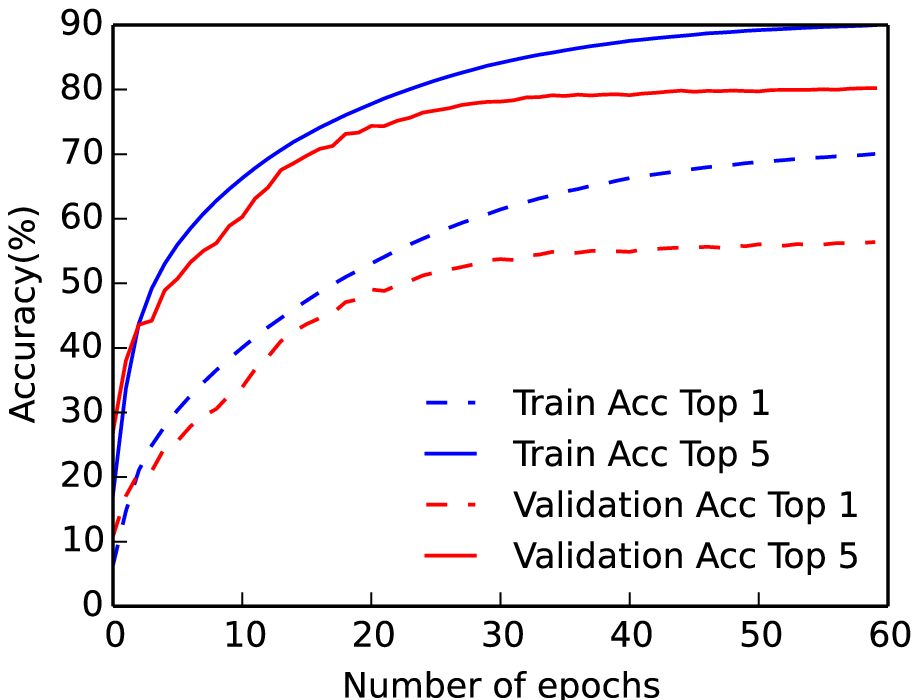}}
		\caption{13 layer net, regular targets}
		\label{fig:f3b}
	\end{subfigure}
		\begin{subfigure}{0.38\linewidth}
		{\includegraphics[width=\linewidth]{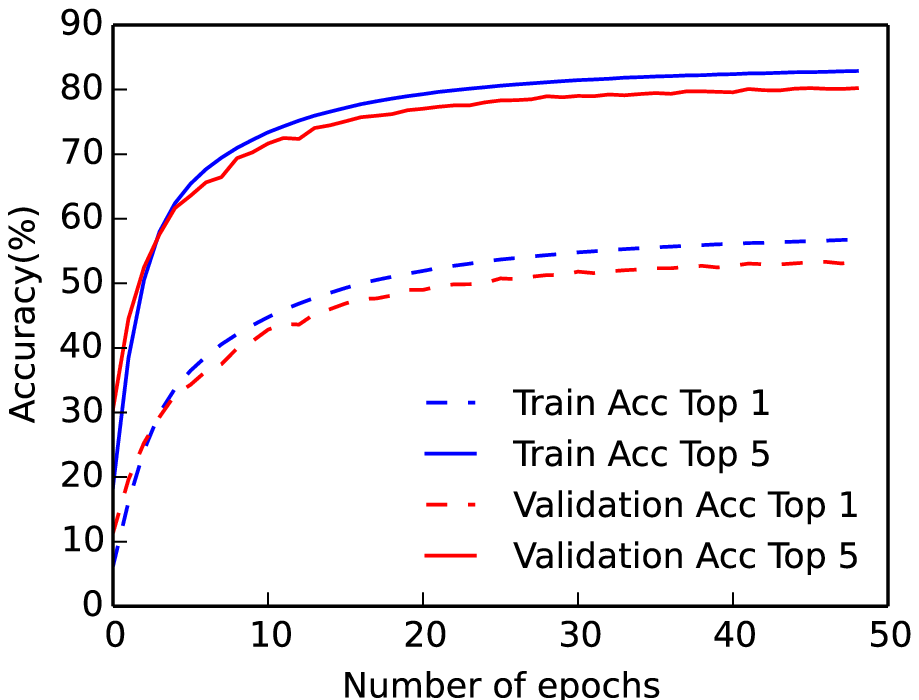}}
		\caption{13 layer net, soft targets}
		\label{fig:f3c}
	\end{subfigure}
		\begin{subfigure}{0.38\linewidth}
		{\includegraphics[width=\linewidth]{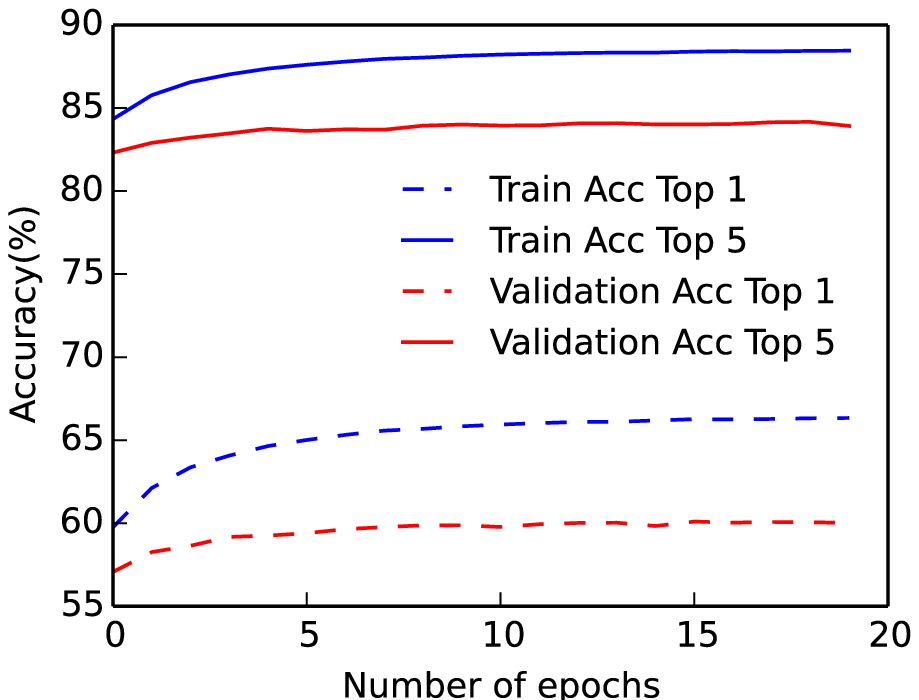}}
		\caption{Post training 13 layer net}
		\label{fig:f3d}
	\end{subfigure}
\end{center}
\caption{Training of high performance networks\\
(a) An BVLC-Alexnet network trained on regular targets. (b) 13 layer network trained with regular targets achieved 80\% top-5 accuracy. (c) 13 layer network trained with soft targets. (d) Fine tuning the 13 layer network with combined soft and regular target achieved 84.1\% top-5 accuracy}
\label{fig:f3}
\end{figure} 

\subsection{Network distillation and avoiding network bottle-neck}
By following aforementioned strategies, we managed to train BVLC-Alexnet \citep{krizhevsky2012imagenet} to 68\% top-5 accuracy rate(Figure \ref{fig:f3a}) slightly better than the one shown in \cite{hubara2016quantized}. Pretraining the network with soft tanh activation gave a bit further improvement and reach 69\% accuracy rate. \\
\begin{table}[h!]
\centering 
\caption{Network architecture}
\label{tab:t1}
\begin{tabular}{|c|c|c|c|c|}
\hline 
\rule[-1ex]{0pt}{2.5ex} Sequence & Type & Channels & Filters & Stride  \\ 
\hline 
\rule[-1ex]{0pt}{2.5ex} 1 & Input & 3 & • & •\\ 
\hline 
\rule[-1ex]{0pt}{2.5ex} 2 & Conv & 128 & 7x7 & 2x2\\ 
\hline 
\rule[-1ex]{0pt}{2.5ex} 3 & Maxpool & 128 & 3x3 & 2x2\\ 
\hline 
\rule[-1ex]{0pt}{2.5ex} 4-6 & Conv & 384 & 3x3 & 1x1\\ 
\hline 
\rule[-1ex]{0pt}{2.5ex} 7 & Maxpool & 384 & 2x2 & 2x2\\ 
\hline 
\rule[-1ex]{0pt}{2.5ex} 8-13 & Conv & 512 & 3x3 & 1x1 \\ 
\hline 
\rule[-1ex]{0pt}{2.5ex} 14 & Maxpool & 512 & 2x2 & 2x2\\ 
\hline 
\rule[-1ex]{0pt}{2.5ex} 15 & Dropout & 512 & • & • \\ 
\hline 
\rule[-1ex]{0pt}{2.5ex} 16 & Dense & 4096 & • & • \\ 
\hline 
\rule[-1ex]{0pt}{2.5ex} 17 & Dense & 1000 & • & • \\ 
\hline 
\rule[-1ex]{0pt}{2.5ex} 18 & Scaling & 1000 & • & •\\ 
\hline 
\rule[-1ex]{0pt}{2.5ex} 19 & Softmax & 1000 & • & •\\ 
\hline 
\end{tabular} 
\end{table}
We also trained a 13 layer network (shown in Table \ref{tab:t1}) and obtained a top-5 accuracy rate of 80.2\%. For this network we used regular weights on last fully connected layer and dropout ratio of 0.2 was used for the dropout layer. Notice wider than usual layers used for the first and second layers, this is designed to avoid information bottleneck associated with binarized networks. The key feature here is that wider layers are only used on early layers and the rest part of the network stay in normal width. This feature is crucial as it maintains very low computing cost brought on by BNNs. 

Further improvement appears hindered by network over-fitting. Adding more layers to the network might further improve the network performance, however such strategy will cut into the computing budget. Since maintaining high efficiency is crucial for real world implementations, we adopted the network distillation \citep{hinton2015distilling}. We used a pre-trained 50 layer residual net\citep{he2015deep} to generate the soft training targets. In agreement with \citep{hinton2015distilling}, we found purely training with soft targets did not improve the network performance(Figure \ref{fig:f3c}). A combined soft and regular targets training significantly improved the network performance to 84.1\%(Figure \ref{fig:f3d}) after the network was pre-trained with soft targets. The network structure used in this part is same as in Table \ref{tab:t1} except no dropout was used and a fixed scaling layer before softmax. Limited by the computing resource available to us, we did not test on bigger networks for better performance.

\section{Discussion and acknowledgement}
\label{others}
Our results show that BNN networks can be trained on ILSVRC2012 data to achieve a good performance with a proper network architectures and training strategies. At this moment the best network we trained is a 13 layers network with 84.1 percent top-5 accuracy rate on a single crop test of validation set. We expect further improvement can be obtained through: 1) testing with multiple scales/crops average as in \citep{he2015deep}, 2) Distill-training with networks of better accuracy, and 3) tuning hyper-parameters beyond a handful of distillation networks we have trained due to very limited computing resources. We are cleaning up the code and will release it soon.

\bibliography{binary}
\bibliographystyle{iclr2017_conference}

\end{document}